\pdfoutput=1

\documentclass[runningheads,a4paper]{llncs}

\usepackage{amssymb}
\usepackage{graphics}
\usepackage{graphicx}
\usepackage{amsmath} 
\usepackage{floatrow}
\usepackage{url}
\usepackage{svg}
\usepackage{gensymb}
\usepackage[ruled,vlined]{algorithm2e}

\urldef{\mailsb}\path|{mdalmasso, agarrell, jimenez, sanfeliu}@iri.upc.edu|
\newcommand{\keywords}[1]{\par\addvspace\baselineskip
\noindent\keywordname\enspace\ignorespaces#1}

\begin{document}

\mainmatter  

\title{Human-robot Collaborative Navigation Search using Social Reward Sources}

\titlerunning{Human-robot Collaborative Search using SRS}

\author{Marc Dalmasso, Ana\'{i}s Garrell, Pablo Jim\'{e}nez and Alberto Sanfeliu}

\institute{ Institut de Rob\`otica i Inform\`atica Industrial (CSIC-UPC) \\
Llorens Artigas 4-6, 08028 Barcelona, Spain\\ \mailsb}
%

\maketitle

\vspace{-5mm}

\begin{abstract} 

This paper proposes a Social Reward Sources (SRS) design for a Human-Robot Collaborative Navigation (HRCN) task: human-robot collaborative search. It is a flexible approach capable of handling the collaborative task, human-robot interaction and environment restrictions, all integrated on a common environment. We modelled task rewards based on unexplored area observability and isolation and evaluated the model through different levels of human-robot communication. The models are validated through quantitative evaluation against both agents' individual performance and qualitative surveying of participants' perception. After that, the three proposed communication levels are compared against each other using the previous metrics.

\end{abstract}

\vspace{-7mm}
\section{Introduction}\label{sec_introduction}
\vspace{-3mm}


On its strife for enhancing life quality, humanity has developed an uncountable number of technologies. Through the years we minimized the effort behind all tasks and automation is a natural consequence of this quest, freeing humans from labour burden and relegating them to supervisory roles. Robotics pursue this ideal, automatic machines capable of physical interaction, motion, learning and adaptation, but some environments offer greater resistance against their intrusion, in particular those populated with humans. Making robots capable of working in social environments is in itself a huge achievement but, despite they may easily outrun humans on some applications, humans still remain as the core experts on many others. The potential of human-robot collaboration cannot be ignored, a coalition capable of exploiting both agents' proficiencies. Achieving effective human-robot collaboration (HRC) is more demanding than previous robotic endeavours. In fact, the main pillars of HRC are: knowledge representation, planning, communication, plan sharing, decision making, agreement and adaptation.


The complexity of social biology is built on instinctual reactions, a feature that inspired the creation of our SRS model summarized in Section III. This model builds an integrated task and world representation, treating action planning and perception as functionally equivalent events, upon which apply global planning methods. Here, we shape this model towards human-robot collaborative search design, exploring the reward space with an adapted motion planning algorithm. Essentially, we define a testbed for model performance evaluation of collaborative object search methods and propose a functional implementation for this task.


In the remainder of the paper, a short review of related work is presented in Section II. Section III briefly defines the concept of ``Social Reward Source (SRS)'' and summarizes the motion planning implementation. Section IV defines the human-robot collaborative search testbed and explains the proposed implementation, Section V specifies the experiments' details and validation metrics and Section VI evaluates the obtained results. Finally, in Section VII, we discuss conclusions and future work.

\section{Related Work}

Human-robot collaboration is a complex and transversal field. To address it, in what follows we encourage a broad view through a brief survey on philosophical and psychological collaboration definitions and fundamentals, a rough discussion about social biology models, as well as briefly appointing some bio-inspired robotics literature, and a review of the current state of the art in human-robot collaboration.

\subsection{Fundamentals on Human Collaboration}

Bratman defined three characteristic features of any shared cooperative activity: mutual responsiveness, commitment to the joint activity and commitment to mutual support \cite{bratman1992}. Sharing a conceptual common ground has huge implications in collaborative tasks \cite{clark1983} and, according to \cite{tomasello2007}, shared intentionality transforms: ``gaze following into joint attention, social manipulation into cooperative communication, group activity into collaboration, and social learning into instructed learning''. Human groups fostering the development of shared task representations are proven to outperform those who don't \cite{vanginkel2009}. In \cite{hommel2001} it is claimed that perceiving and action planning are functionally equivalent: internally representing external events. It is interesting to note that humans are capable of representing robot actions in a similar manner \cite{wykowska2014}.

\subsection{Biology Inspiration}

Interactions based on long-lasting chemical marks that trigger instinctual reactions in other individuals cached our attention due to their low cognitive requirements and their broad transversality along species. Especially, the usage of such channels by social insects \cite{vander1998}. We found special potential in ants communication, illustrated by their job-specific trail marking pheromones, combining both positive and negative feedbacks, capable of signalling long and short term attractive paths and temporal avoidance of such \cite{jackson2006}. From all insect-inspired models, we may highlight those making use of virtual pheromones \cite{brambilla2013,narzt2010,susnea2009}.

\subsection{Human-Robot Collaboration}

When facing human-robot joint action, it is of utmost interest to analyze disciplines as human-human joint action and connect them to the human-robot joint action case \cite{hoffman2004,clodic2017}. Theory of mind approaches take importance as we try to model the knowledge of the robot: \cite{devin2016} estimates and maintains mental states of other agents reducing the unnecessary information given to the human and \cite{lemaignan2017} claims to have built a cognitive robot to successfully share collaborative spaces and tasks. Moreover, Roncone's proposal \cite{roncone2017} is able to autonomously reason about the problem of allocating specific subtasks to either the robot or its human partner. Many interesting efforts approached physical human-robot collaboration (PHRC), 
a detailed survey of this field can be found in \cite{ajoudani2018}.


\subsubsection{Metrics for HRC}

It has become necessary to quantitatively analyze the performance of the heterogeneous teams to enable comparison between different team configurations.
Recently, \cite{hoffman2019} reviewed present subjective and objective fluency metrics. He suggests to carefully observe objective metrics dynamic behaviour, given their variability, and studies their correlation with subjective metrics.

\subsubsection{Human-Robot Collaborative Navigation}

One of the first faced human-robot collaborative challenges was side-by-side navigation \cite{morales2014,nakazawa2015}. In parallel, \cite{ferrer2013a,ferrer2017,garrell2017b} approached this challenge through Social Force Model (SFM) methods and, in another context, \cite{nguyen2016,jayawardena2018} presented methods for side-by-side wheelchair navigation. Alternative approaches to HRCN include co-driving, as the collaborative teleoperation of a robot through dialogue \cite{fong2003} or the collaborative control of wheelchair \cite{carlson2012}. They are the first steps into collaborative models, but they are task-focused thus can't be extended to other applications. We pursue a flexible model capable of representing multiple tasks and conveying such representations to the human. 

\section{Social Reward Sources}

The SRS model is based on two primary instincts, attraction and repulsion, or in other terms, positive and negative perception of one's state. We can find many literature applications based on reward function definitions in the literature, but the SRS model aims not to describe the final reward, but to model its sources. It can be seen as a generative model framework, as it describes sources properties and dynamics affecting the final global reward. As introduced before, such model is inspired in social biology mechanisms, as in the case of virtual pheromones. Logically it easily extends to repulsion over personal space invasion, but these social reward sources may encode higher level abstractions. This includes, for example, the satisfaction over fulfilling a task, the propensity to follow someone's instructions or the discomfort felt when obstructing other people actions, as when standing in front of a person trying to take a photograph. This model aims to integrate and unify world and task representations, human-robot communication and human social or profiling preferences in a unique interrelated framework.

\begin{table}[htb]
\centering
\resizebox{\textwidth}{!}{%
\caption{$SRS$ implementation over sampling methods}
\begin{tabular}{|c|c|}

\hline
\bfseries Path Generation & \bfseries Path Selection \\
\hline
\rule{0pt}{0ex}&\\  
$ C^{gen}_n = C^{gen,cm}_n + C^{gen,cs}_n $
&
$ C^{sel}_n = C^{sel,cm}_n + C^{sel,cs}_n + C^{sel,f}_n $
\\
\rule{0pt}{0ex}&\\  
$ C^{gen,cm}_n = $
&
$ C^{sel,cm}_n = $
\\
$ \sum_{i}^{P_n} \Big( \sum_j^{S^{gen,cm}} (s_j(i) \cdot d_{i-1,i}) + c^{conn}_{i-1,i} \Big) $
&
$ \sum_{i}^{P_n} \Big( \sum_j^{S^{sel,cm}} (s_j(i) \cdot d_{i-1,i}) + c^{conn}_{i-1,i} \Big) $
\\
\rule{0pt}{0ex}&\\  
$ C^{gen,cs}_i = \sum_j^{S^{gen,cs}}(\max_{i}^{P_n} \{ s_j(i)\}) $
&
$ C^{sel,cs}_i = \sum_j^{S^{sel,cs}}(\max_{i}^{P_n} \{ s_j(i)\}) $
\\
\rule{0pt}{0ex}&\\  
&
$ C^{sel,f}_i = \sum_j^{S^{sel,f}} s_j(i) $
\\
\rule{0pt}{0ex}&\\  
\hline
\multicolumn{2}{|c|}{ } \\
\multicolumn{2}{|c|}{Where $C^{p,k}_n$ is the cost contribution to the path $P_n$ ending at node $n$} \\
\multicolumn{2}{|c|}{of $S^{p,k}$, the set of sources $s$ of nature $k$ (possibly being cumulative $cm$,} \\
\multicolumn{2}{|c|}{consumable $cs$ or final $f$) and application policy $p$ (path generation $gen$} \\
\multicolumn{2}{|c|}{and/or path selection $sel$). Also, $c^{conn}_{a,b}$ and $d_{a,b}$ denote respectively the} \\
\multicolumn{2}{|c|}{connection cost and distance from $a$ to $b$, being the latter in the dimen-} \\
\multicolumn{2}{|c|}{sional magnitude over which cumulative cost densities are defined.} \\
\multicolumn{2}{|c|}{ } \\
\hline

\end{tabular}
\label{equationrrt}
}
\end{table}

\begin{multicols}{2}

\begin{algorithm}[H]
\vspace{0.2cm}
\SetAlgoLined
 T $\leftarrow$ get\_tree\_origins(S)\;
 initialization\;
 \While{n $<$ nodes or t $<$ time}{
  expand\_rrt($S^{gen}$)\;
  }
  
 eval\_rrt($S^{sel}$)
 \caption{$S-RRT^{\star}$}
\end{algorithm}

\begin{algorithm}[H]
\SetAlgoLined
  node $\leftarrow$ sample\_node()\;
  \For{$t \in T$}{
    t $\rightarrow$ addNode(added)\;
    \If{added}{
    node $\rightarrow$ set\_cost($C^{gen}_n$)\;
    t $\rightarrow$ rewire(node)\;
    \If{not first added}{merge\_trees()\;}
  }
  }
 \caption{expand\_rrt}
\end{algorithm}
 
\end{multicols}

Ultimately, the only requirement for a reward source is to correctly generate a reward function defined along all the search space. Nevertheless, consistent spatial properties of humans' world abstractions, such as objects, rooms or demonstrative references, inspired setting a spatial interpretation for such sources. Due to these and other functional and dynamic considerations, each source is defined by the following properties: type (repulsive or attractive), application policy (i.e. path generation and/or path selection), nature (cumulative, consumable or final), model (i.e. standard decay, as Gaussian or power function models, or complex definitions, as the graph-built observability presented in this paper), shape and dynamics (movement).

\begin{figure}[tb]
    \centering
    \includegraphics[height=0.23\textwidth]{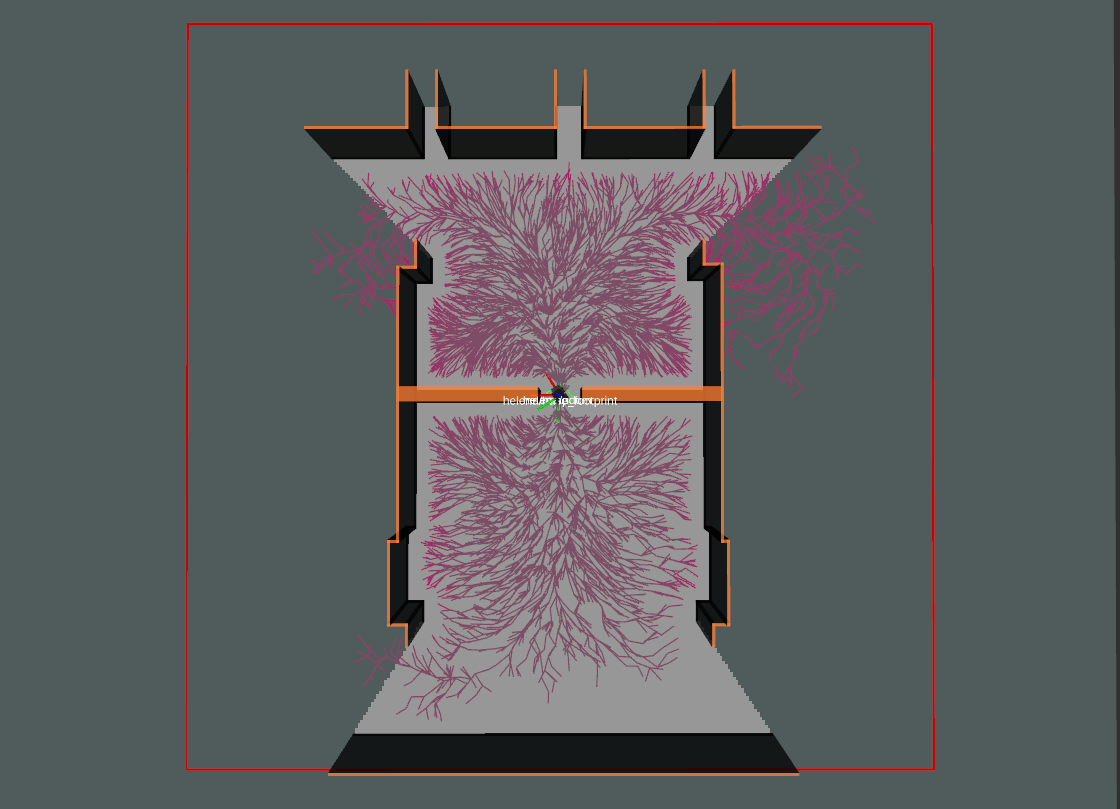}
    \includegraphics[height=0.23\textwidth]{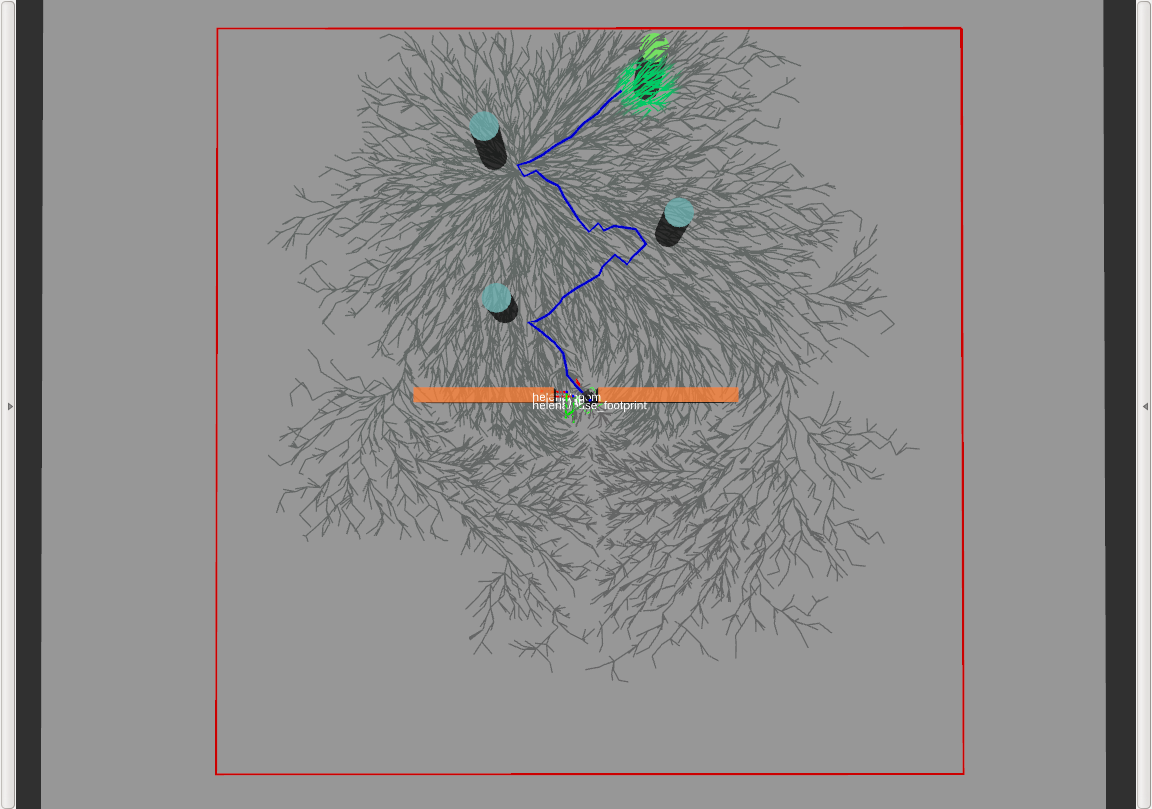}
    \includegraphics[height=0.23\textwidth]{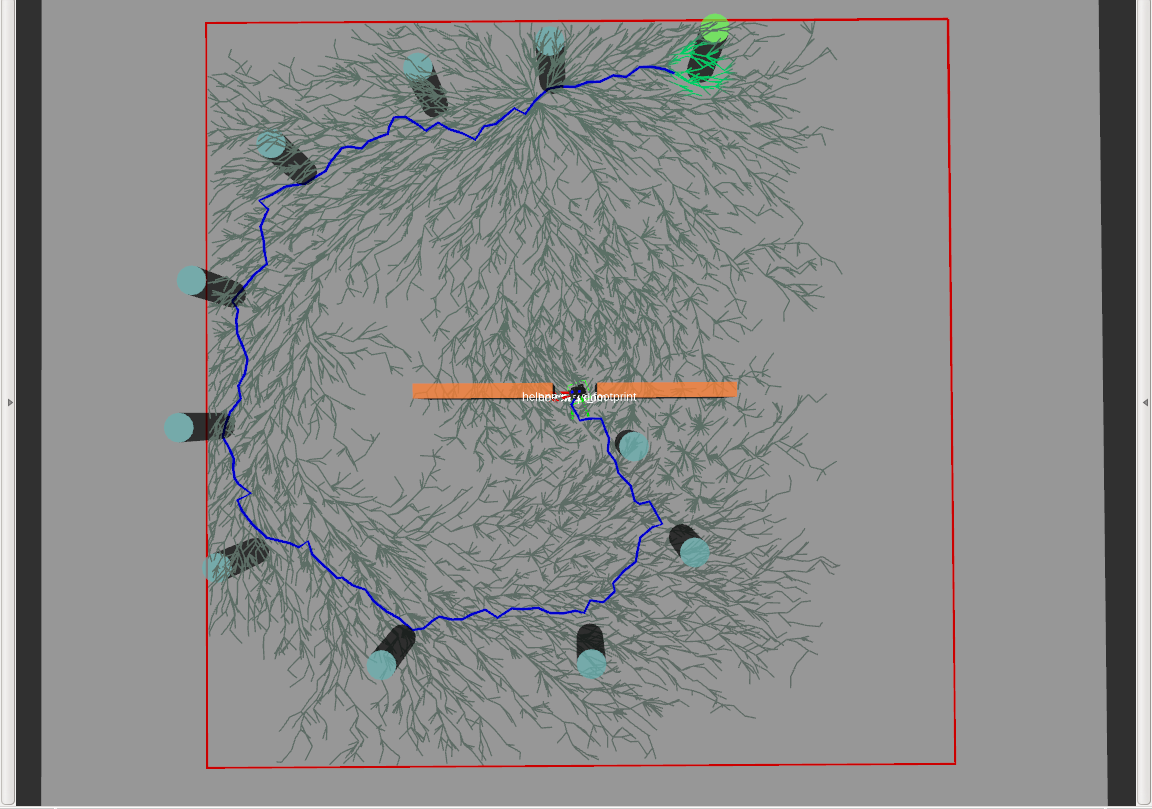}
    \caption{{\bf Social Reward Sources.} $S-RRT^{\star}$ expansion and path selection for, from left to right: a) A hallway environment, b and c) two consumable reward sources distributions ending on a final reword source.}
\label{implmentationrrt}
\end{figure}

When exploring the generated rewards, they can be mirrored to understand them as costs. Essentially, any negative reward can be seen as the perceived cost of receiving it, while a positive one can be modelled as a negative cost. Any motion planning algorithm that takes these costs in consideration is a suitable search engine for exploring space and computing a path. Here we have adapted the well-known $RRT^{\star}$ algorithm (Rapidly exploring Random Trees) due to its computational efficiency. We call the resulting algorithm $S-RRT^{\star}$, where $S$ stands for ``Sourced'' emphasizing the usage of reward's sources models. The computation of the relevant costs within $S-RRT^{\star}$ is summarized in Table \ref{equationrrt}. Moreover, some applications of the method are shown in Fig. \ref{implmentationrrt}.

\section{Human-Robot Collaborative Search}

The main issue in collaborative search is to share the exploration progress with each other. Several approaches to map sharing have been published for multi-robot collaboration, but such approaches are not suitable for a human's mental map. Instead, humans actively do infer others' knowledge from their actions while, at the same time, they expect to be inferred themselves. Only in doubtful situations, they do resort to specific task-related active communication. As a matter of fact, humans are experts in social and navigation tasks, and thus interacting with a robot can easily become boring or burdening.

The collaborative search testbed has been defined as follows: both the human and the robot know the map of the search zone beforehand and the searched object can be in any place of the unexplored zone with uniform probability. The task ends when either one of the agents finds the object. In simulation experiments, exploration progress is shown to the human to avoid misestimation of the observed zone. Communication between the robot and the human can be arbitrarily extended to enhance joint activity performance or fluency.

\begin{figure}[tb]
    \centering
    \includegraphics[width=0.24\textwidth]{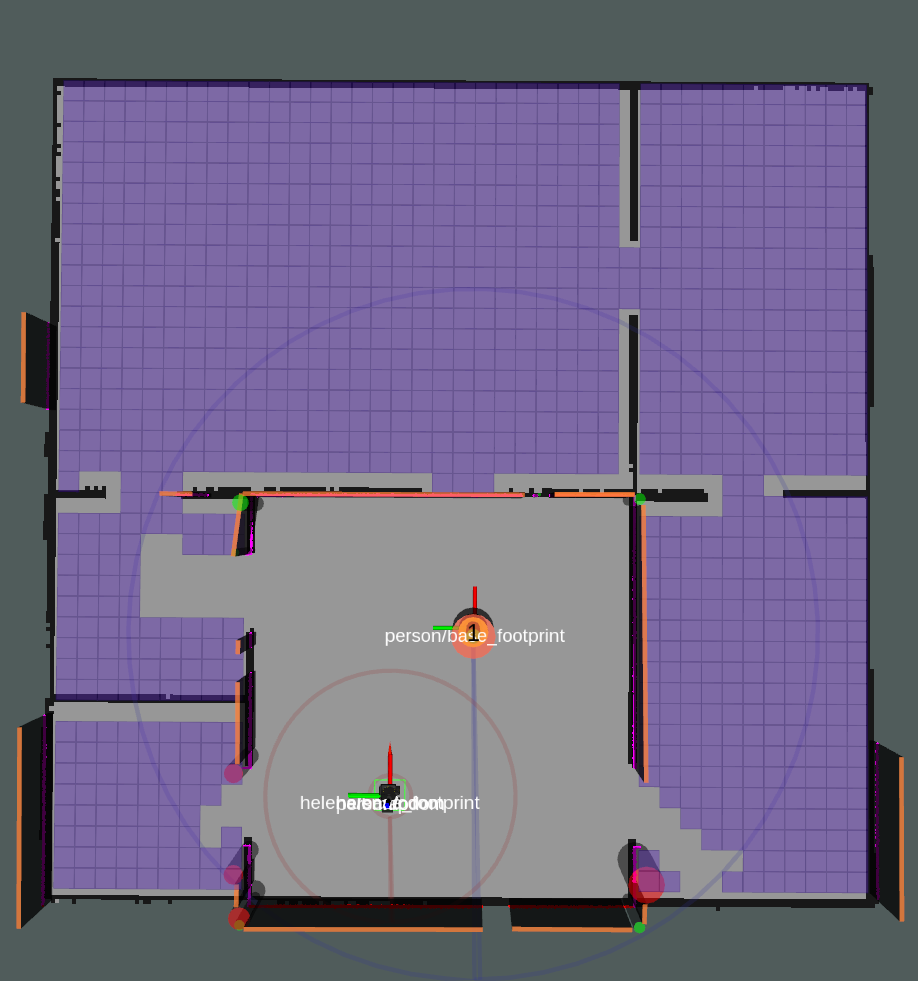}
    \includegraphics[width=0.24\textwidth]{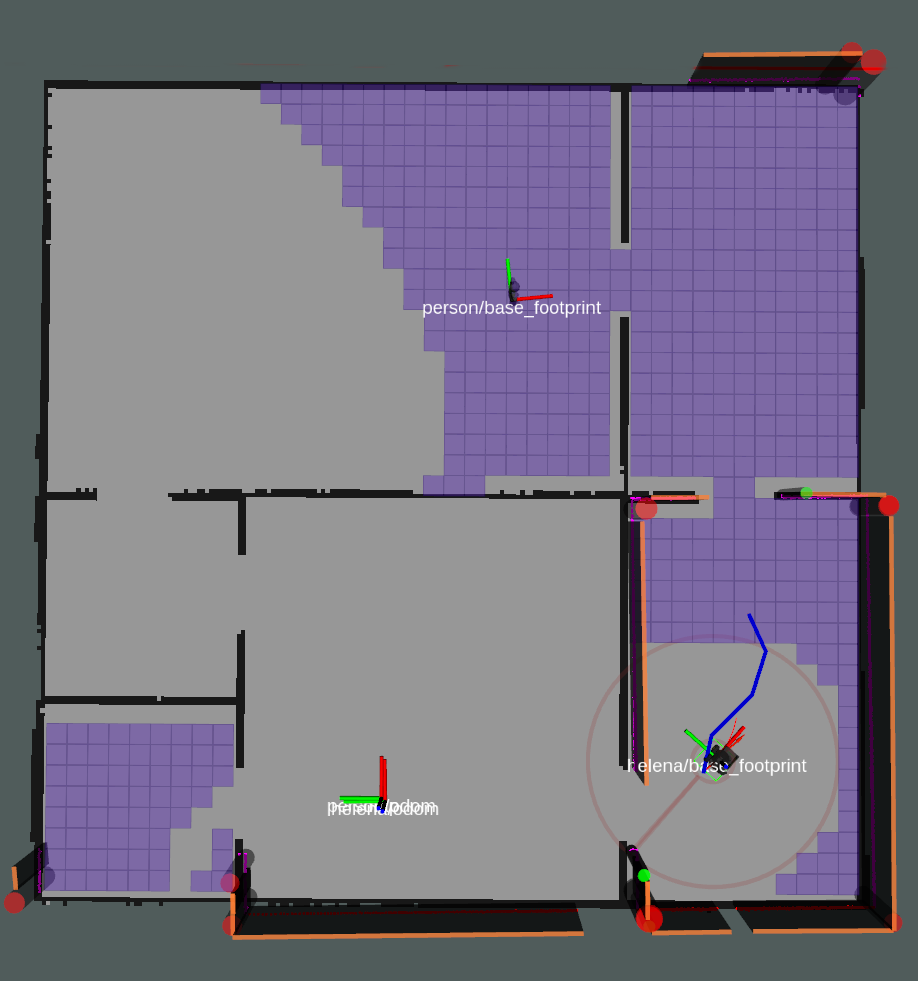}
    \includegraphics[width=0.24\textwidth]{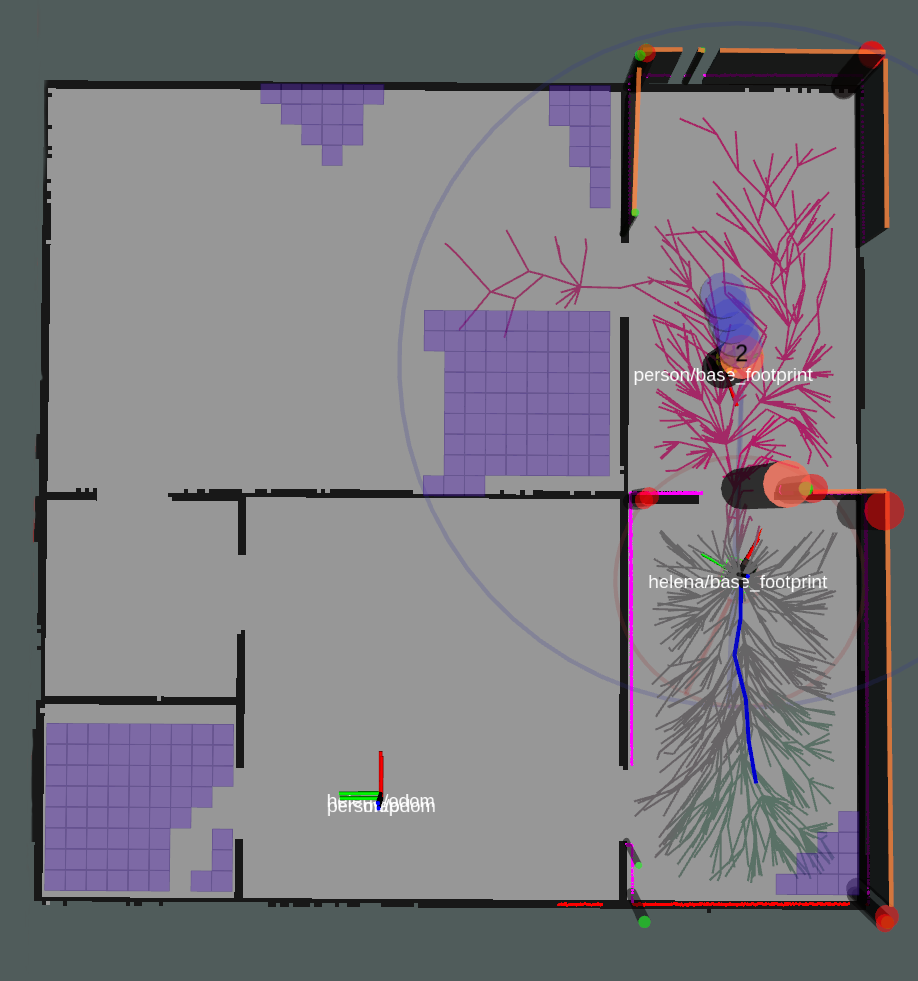}
    \includegraphics[width=0.24\textwidth]{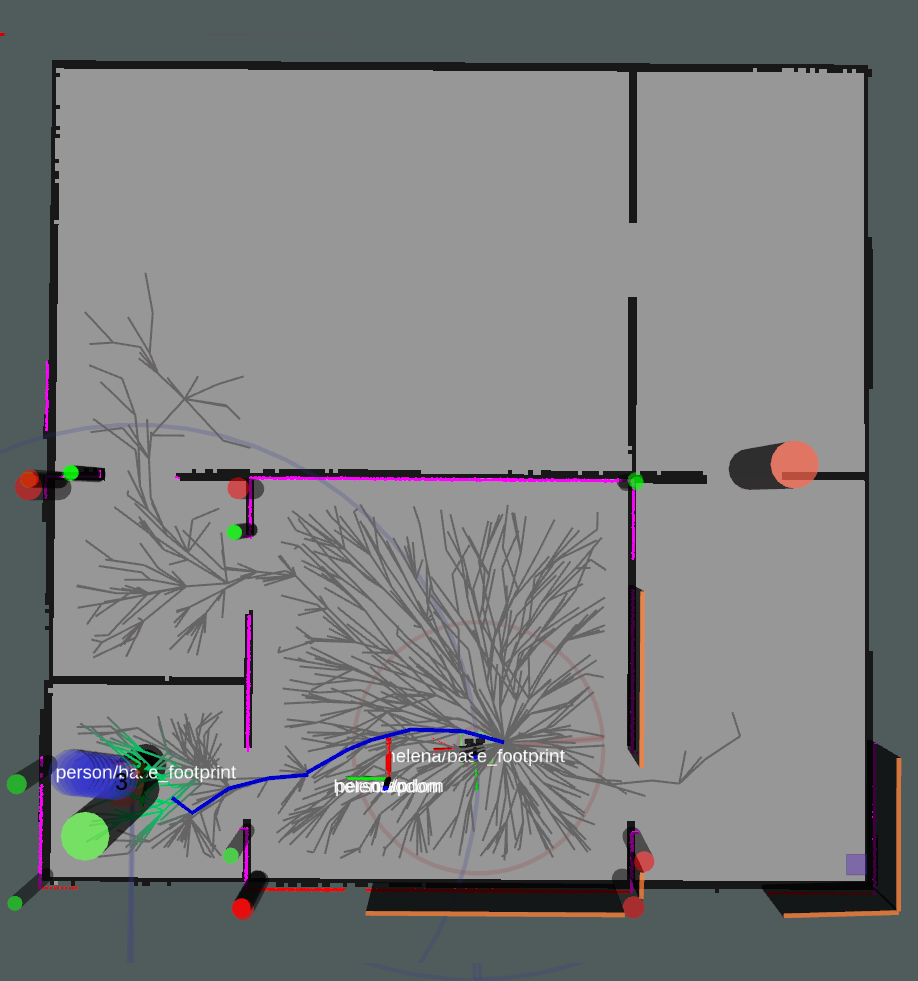}
    \caption{{\bf Collaborative Search Testbed.} From left to right: a) The robot infers the unexplored zone from its detection range (red circle) and the person's (blue circle). b) People detection is impossible when the person is out of sight, hence no inference is done. c) The person indicates the robot to avoid searching through that zone, as either it is already explored or the person will do it on their own. d) The person finds the object, thus indicates the robot to come.}
\label{collabsearchtestbed}
\end{figure}

In our implementation, both human and robot detection capabilities are defined as radial distances, their field of view is assumed of $360\degree$ and no detection uncertainty is considered, as observable in (Fig. \ref{collabsearchtestbed}.a). The human is detected and tracked by the robot through 2D laser sensors and the robot knowledge of his or her contribution to the exploration is inferred accordingly (fig \ref{collabsearchtestbed}.b). Two communication examples using our model are shown in Fig. \ref{collabsearchtestbed}.c and \ref{collabsearchtestbed}.d, being respectively to ``avoid going through a zone'' or to ``go to one place''.

\begin{figure}[t]
    \centering
    \includegraphics[width=0.327\textwidth]{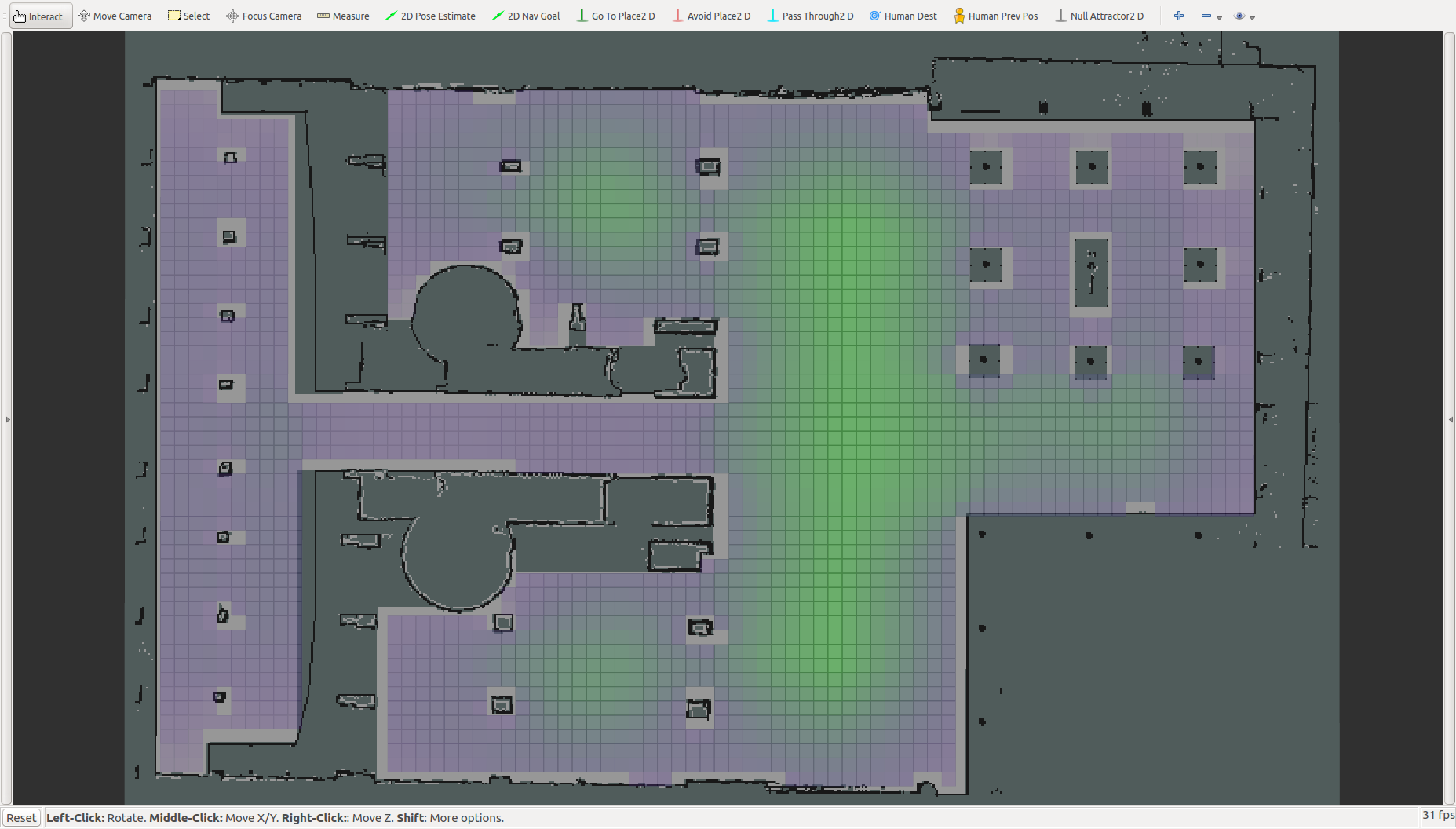}
    \includegraphics[width=0.327\textwidth]{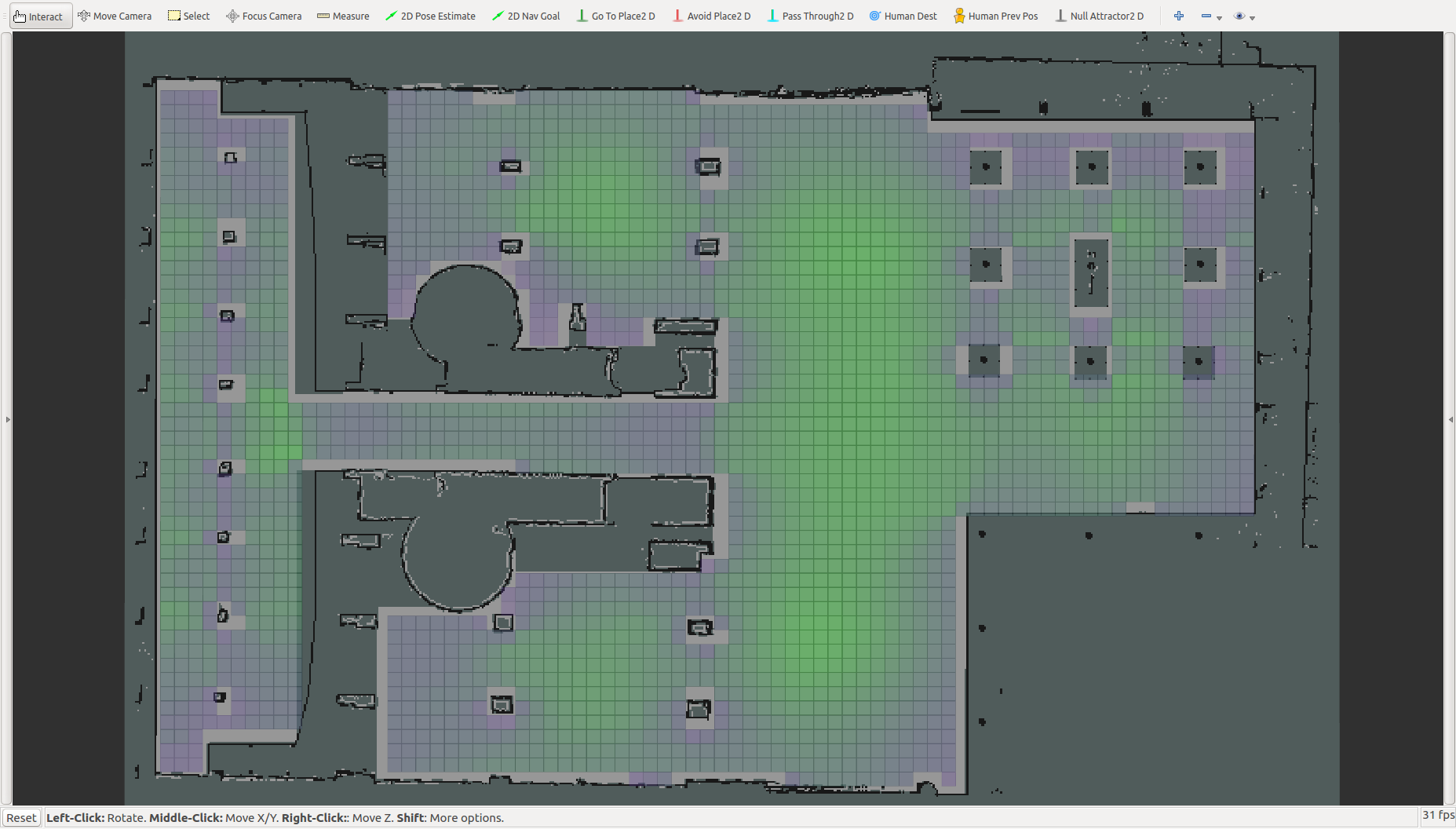}
    \includegraphics[width=0.327\textwidth]{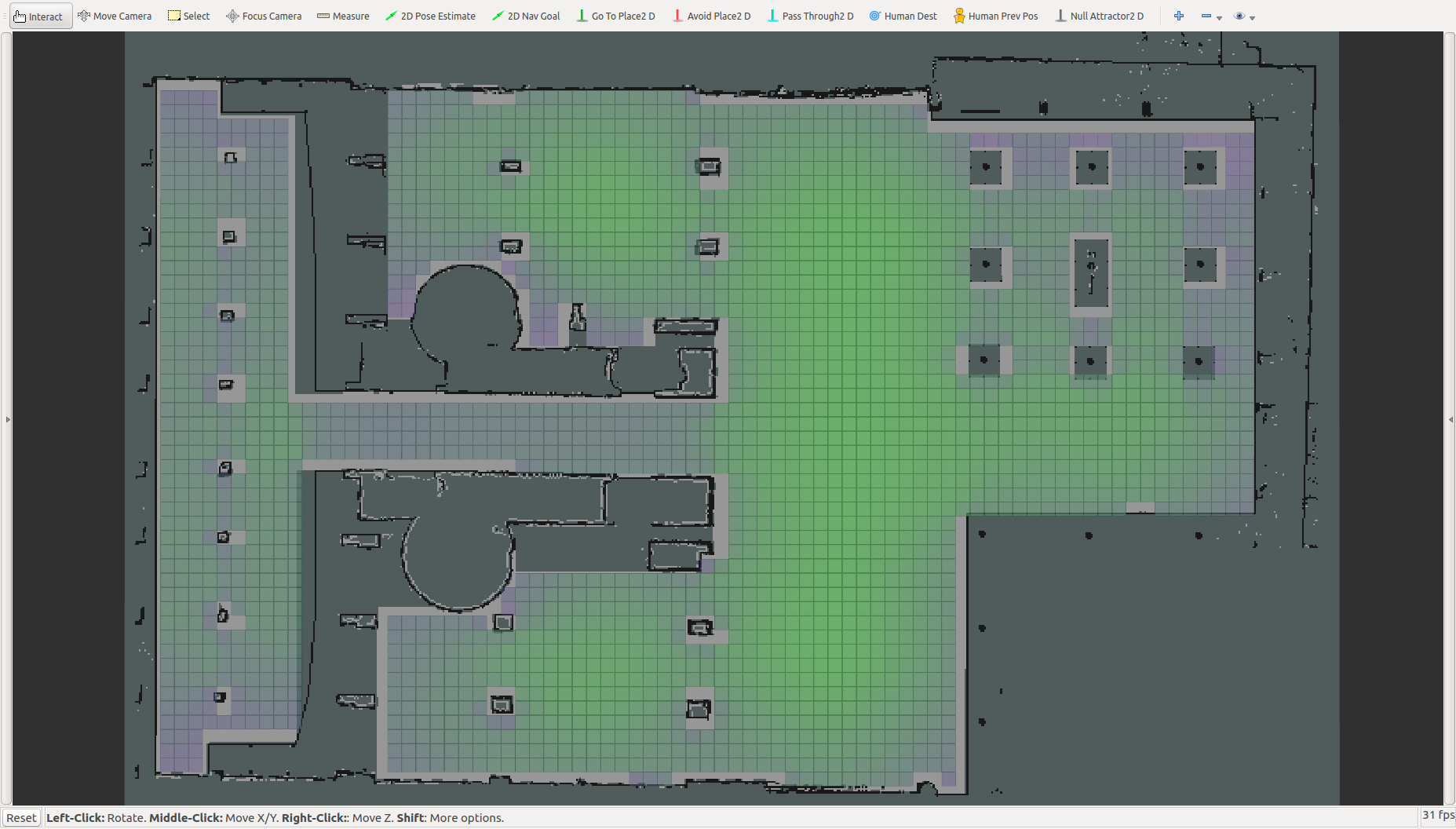}
    \caption{{\bf Collaborative Search Source.} From left to right: a) Observability score of the current map exploration. b) Isolation Score of the current map exploration. c) Search reward generated by the collaborative search source for $w_o = 0.8$ and $w_i = 0.2$, the values found to work best.}
\label{collabsearchattr}
\end{figure}

Aiming at a specific model for the object search task, we discretised the explorable area and built an observability graph. We model the belief of seeing the searched object from one place as the observable unexplored zone from it (Fig. \ref{collabsearchattr}.a). Similarly, we model the search isolation of one place as the inverse of the mean observability of the observable area from this point (Fig. \ref{collabsearchattr}.b). Both values are normalized and merged in a weighted sum, the second being added to tune robot eagerness to clear neighbouring non-observed isolated zones before addressing bigger zones. This combination is normalized and weighted on a logarithmic scale to construct the final search reward shown in Fig. \ref{collabsearchattr}.c.

\begin{subequations}
\begin{equation}
    O(p) = \sum_{p_i}^{obs(p)} B(p_i)
\end{equation}
\begin{equation}
    I(p) = \Big( obs(p) / \sum_{p_i}^{obs(p)} O(p_i) \Big) \cdot O(p)
\end{equation}
\vspace{1mm}
\begin{equation}
    S(p) = O(p) / \max_{p_i} (O(p_i)) \cdot w_o + I(p) / \max_{p_i} (I(p_i)) \cdot w_i
\end{equation}
\vspace{2mm}
\begin{equation}
    R(p) = \log (S(p)/\max_{p_i} (S(p_i)) + 1)
\end{equation}
\end{subequations}
\vspace{2mm}

given $B(p)$ is the prior probability of the object being at location $p$ and $obs(p)$ is the observable zone by the robot from $p$. $O(p)$, $I(p)$ and $S(p)$ are respectively the observability, isolation and search scores of location $p$. $w_o$ and $w_i$ are the tuned weight values for observability and isolation, $R(p)$ is the normalized reward nominal value of $p$. The search social reward source is of final nature and applied in the path selection phase. 

\section{Validation}

\begin{figure}[t]
    \centering
    \includegraphics[width=0.327\textwidth]{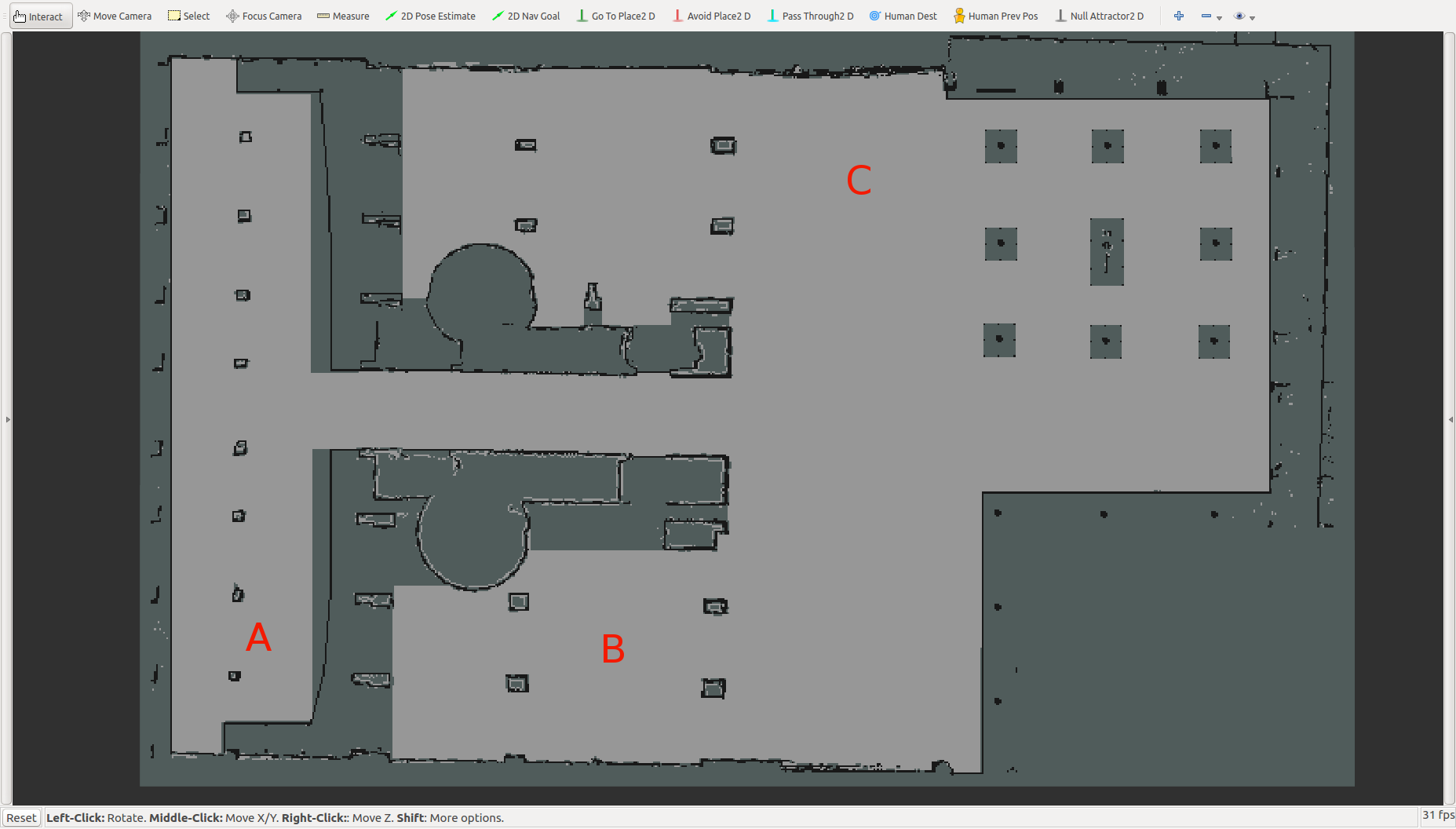}
    \includegraphics[width=0.327\textwidth]{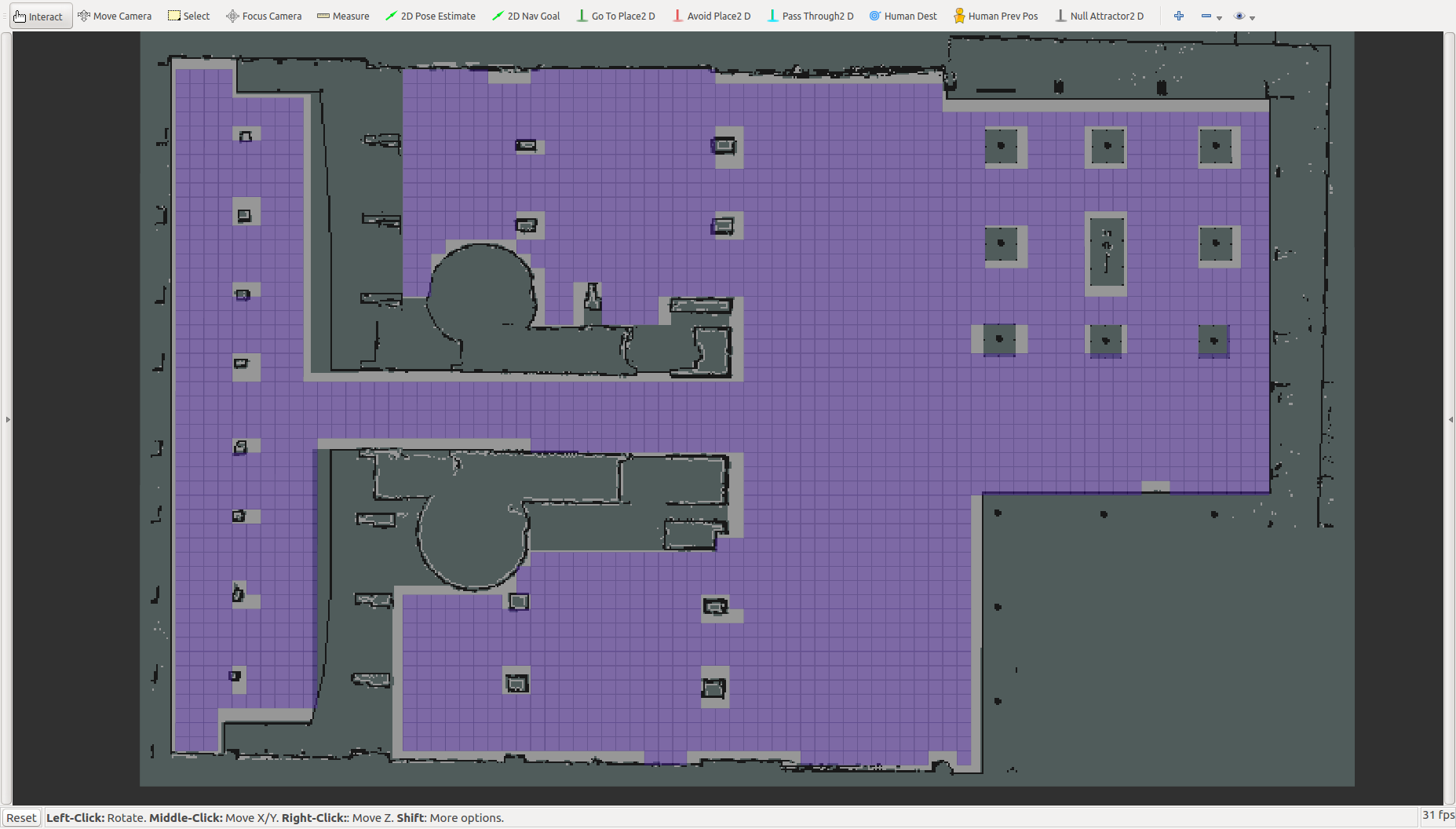}
    \includegraphics[width=0.327\textwidth]{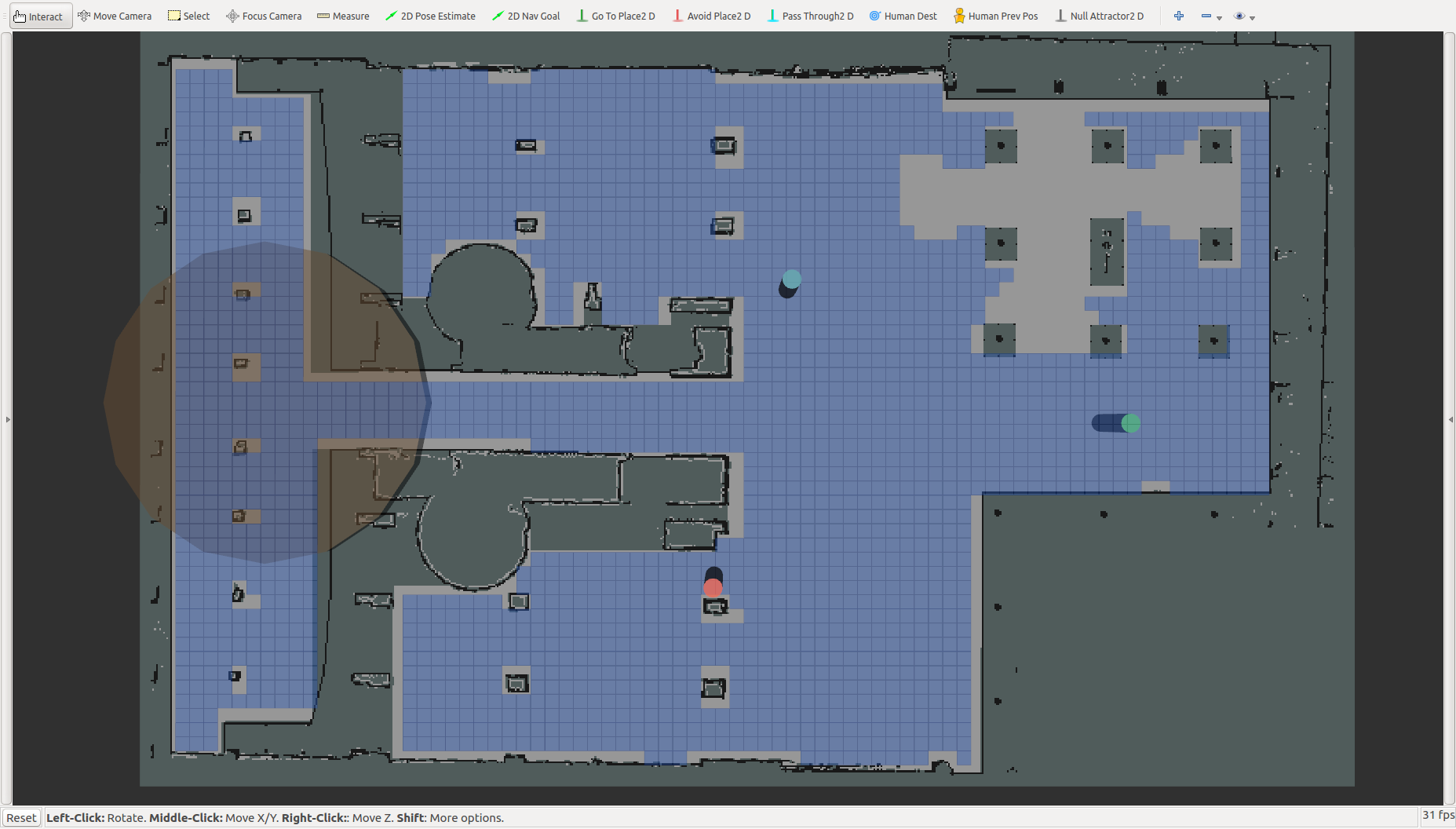}
    \caption{{\bf Collaborative Search Experiments.} From left to right: a) BRL map. b) Search space discretisation. c) Robot perceived exploration progress and visual feedback of the communication instructions given to the human: ``go to this place'' (green cylinder), ``pas through this place'' (blue cylinder), ``avoid this place'' (red cylinder), ``I'm going to this place'' (brown area) and ``I've already been here'' (perceived explored area at the top right zone of the map.}
\label{collabsearchexp}
\end{figure}

To validate our model, we chose the BRL map from the \textit{Barcelona Robot Lab Dataset}\footnote{\url{http://www.iri.upc.edu/research/webprojects/pau/datasets/BRL/}} and defined three different origins to begin the search (Fig. \ref{collabsearchexp}.a). The considered explorable area is discretised and shown in Fig. \ref{collabsearchexp}.b and all objects in the scene are assumed to block both the view of the robot and the human.

First, we tested human and robot individual search performance to establish a baseline. After that, we tested the human-robot collaborative search model through three different communication levels. In the first one, the human was only able to see the exploration progress and the robot location. In the second approach, the robot showed the human his perceived exploration progress and his current planned path. During the third experiment, the human was able to communicate with the robot through 5 instructions (Fig. \ref{collabsearchexp}.c): three general instructions (``go to this place'', ``pass through this place'' and ``avoid this place'') and 2 task-related informative messages (``I'm going to this place'' and ``I've already been here'').

A total of 12 volunteers participated in the experiment, with ages between $15$ and $34$ (mean: $26.42$ std: $5.23$). On a scale of 1 (None) to 7 (Expert) their average self-evaluated knowledge in robotics was $4.83$ (std: $1.64$). No one had any experience using the framework, neither were they given the chance to practice. Each of them participated in three or four of experiment setups involving humans, doing 3 or 6 episodes on each one equally distributed among the different origins. Additionally, participants were surveyed after each communication level setup whether they perceived robot plan as efficient and how much did they change their plans due to the robot actions. Both questions were answered on a linear scale from 1 (not at all) to 7 (completely).

During all the experiments, both the speed of the robot and the human were limited. The robot was able to move at a maximum linear speed of $0.7$ m/s, being it the nominal maximum velocity of the real robot, a luggage transporter mounted on a \textit{Pioneer P3-DX} base. The human maximum velocity was limited to $1$ m/s and it's movement controlled through a \textit{PlayStation 3 Dualshock 3 Wireless Controller}. The final mean speeds of the human and the robot during the simulations were $0.83$ m/s and $0.53$ m/s.





\section{Results \& Discussion}

\begin{figure}
    \centering
    \includegraphics[width=1.0\textwidth]{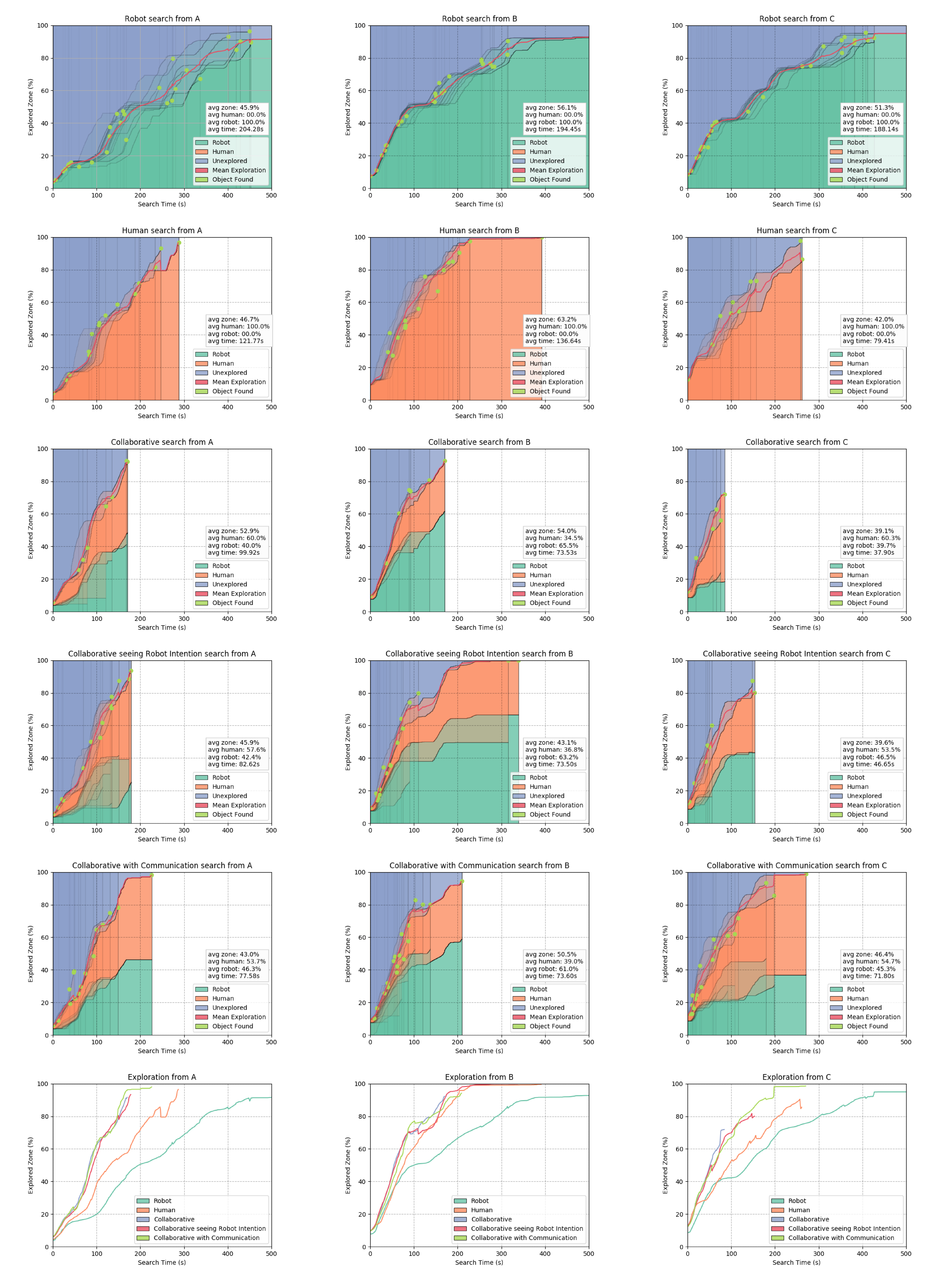}
    
    \caption{{\bf Human-Robot Collaborative Search Experiments.} From left to right: episodes beginning at origins A, B and C. From top to bottom: robot individual search, human individual search, collaborative search, collaborative search seeing robot intention, collaborative search including human to robot communication and comparison between the 5 setups, both in performance and concurrent activity.}
\label{collabsearchdata}
\end{figure}

A complete plot of the collaborative search dataset is shown in Fig. \ref{collabsearchdata}. Here, we can observe origin selection has a strong effect on the search progress dynamics. Although the robot is slower, we can observe correlations between the human and the robot search progress shape, suggesting their search policies are alike.

Robot behaviour consistently shows greater variability when beginning in origin A until the last collaborative setup. Such variability presumably appears due to the presence of two major bifurcations. Consistency in the collaborative search with communication dataset suggests human users either instructed the robot where to go or implicitly conditioned its choice by providing it with information. As a matter of fact, all the participants preferred the robot to take the hallway while they explored the remaining area in their side. Moreover, most of them enforced this behaviour through direct orders, while the usage of the task-related informative messages was relegated only to the right part of the map.

Episodes beginning in B have the biggest robot contribution. In this origin, after exploring the little zone at the left, both the robot and the user are enforced to take the same direction. That obstructed searching in parallel. Except for late-stage search progress when beginning in this origin, all three collaborative models surpassed both the individual human and robot baselines. In terms of search progress, however, neither of the three is significantly better than the others. We judge that adaptation capabilities of the human, as well as its superior movement capabilities, made up for the lack of communication. Besides, the information given to the human on the first collaborative setup might be too extense, this encourages further experiments conveying less information to the human.

\begin{figure}[tb]
    \centering
    \includegraphics[width=0.4\textwidth]{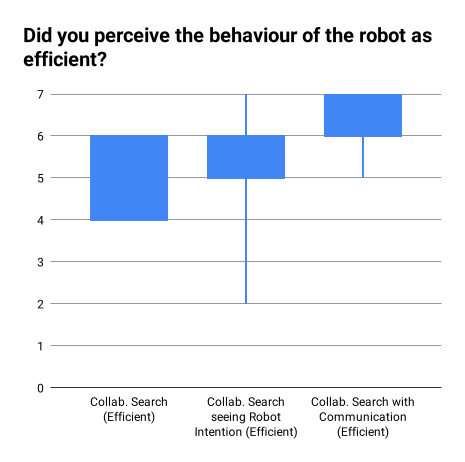}
    \includegraphics[width=0.4\textwidth]{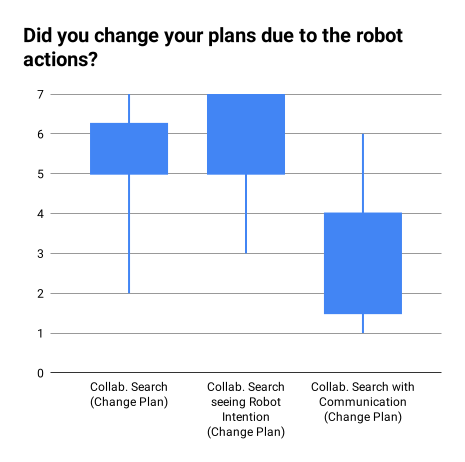}
    \caption{{\bf Collaborative Search Survey}}
\label{collabsearchsurvey}
\end{figure}

Human subjective perception of the task, however, does change between the three collaborative setups. Including human to robot communication seems to increase the human perception of the robot efficiency and greatly decrease situations where the human is forced to adapt to the robot. Differences between the other two models are less clear. Even though in the second one the human had a broader perception of the robot intention, this might have enhanced conflict situations between the human-perceived robot plan and their own. Results of the survey are represented in Fig. \ref{collabsearchsurvey}.

\section{Conclusions}

In this paper, we presented a complete human-robot collaborative navigation task implementation in the SRS framework, which is proven to outperform the individual search baseline. Moreover, human to robot communication is proven to have a major impact in human perception of human-robot collaborative tasks, while performance might not be significantly affected in simple setting due to the human adaptation capabilities.

We aimed to adapt fluency metrics analyzed in \cite{hoffman2019}. However, their dynamics didn't seem to correlate with the results obtained in the qualitative survey, which may suggest the need to search for other quantitative metrics. Moreover, to do so we identified actual progress in the exploration, as identifying all goal-driven movement would result in the trivial case of not having idle time in any agent.

This is a first approach tackling task-oriented explicit collaborative navigation. In future work, we will expand this model to include theory of mind knowledge models, shared planning and agreement mechanisms.

\vspace{-1em}

\section{Acknowledgements}

Work supported under projects ColRobTransp (DPI2016-78957-RAEI/FEDER EU), TERRINet (H2020-INFRAIA-2017-1-two-stage-730994) and by the Spanish State Research Agency through the Maria de Maeztu Seal of Excellence to IRI (MDM-2016-0656).

\bibliographystyle{splncs03.bst}

\end{document}